\documentclass[10pt,twocolumn,letterpaper]{article}

\usepackage{cvpr}
\usepackage{times}
\usepackage{epsfig}
\usepackage{graphicx}
\usepackage{amsmath}
\usepackage{amssymb}

\usepackage{multirow}
\usepackage{verbatim}
\usepackage{color}
\usepackage{float}
\usepackage{enumitem}
\usepackage{booktabs}
\usepackage{tabulary,multirow,overpic,xcolor}
\usepackage{caption}
\usepackage{subcaption}
\usepackage[font={small}]{caption}
\usepackage[title]{appendix}

\usepackage[british,UKenglish,USenglish,english,american]{babel}

\newcolumntype{x}[1]{>{\centering\arraybackslash}p{#1pt}}

\newcommand{\app}{\raise.17ex\hbox{$\scriptstyle\sim$}}

\newcolumntype{x}[1]{>{\centering\arraybackslash}p{#1pt}}

\newlength\savewidth\newcommand\shline{\noalign{\global\savewidth\arrayrulewidth
  \global\arrayrulewidth 1pt}\hline\noalign{\global\arrayrulewidth\savewidth}}
\newcommand{\tablestyle}[2]{\setlength{\tabcolsep}{#1}\renewcommand{\arraystretch}{#2}\centering\footnotesize}

\newcommand{\customfootnotetext}[2]{{
  \renewcommand{\thefootnote}{#1}
  \footnotetext[0]{#2}}}

\setlist[itemize]{leftmargin=*}

\usepackage[pagebackref=true,breaklinks=true,letterpaper=true,colorlinks,bookmarks=false]{hyperref}

\cvprfinalcopy 

\def\cvprPaperID{2746} 

\ifcvprfinal\fi
\begin{document}

\title{Learning Correspondence from the Cycle-consistency of Time}

\author{
Xiaolong Wang\textsuperscript{*} \\
Carnegie Mellon University
\\
{\tt\small xiaolonw@cs.cmu.edu}
\and
Allan Jabri\textsuperscript{*} \\
UC Berkeley
\\
{\tt\small ajabri@eecs.berkeley.edu}
\and
Alexei A. Efros\\
UC Berkeley
\\
{\tt\small efros@eecs.berkeley.edu}
}

\twocolumn[{%
\vspace{-1em}
\maketitle
\vspace{-1em}

\begin{center}
    \centering 
    
    \vspace{-0.3in}
    \includegraphics[width=\linewidth]{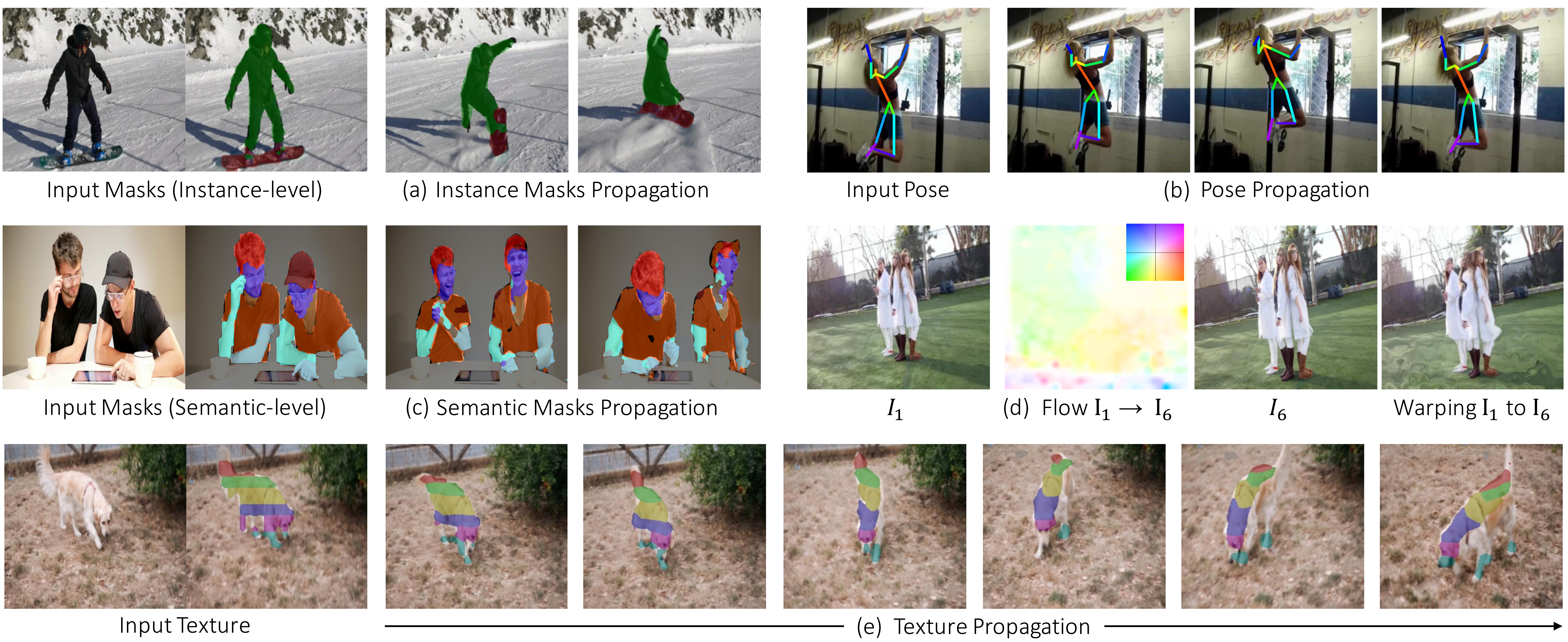}
    \vspace{-0.25in}
    \captionof{figure}{
    We  propose to learn a representation for visual correspondence from raw video. Without any fine-tuning, the acquired representation generalizes to various tasks involving visual correspondence, allowing for propagation of: (a) Multiple Instance Masks; (b) Pose; (c) Semantic Masks; (d) Long-Range Optical Flow; (e) Texture. }
    \label{teaser}
\end{center}
}] 
\customfootnotetext{*}{Equal contribution.}

\begin{abstract}
\vspace{-3mm}
We introduce a self-supervised method for learning visual correspondence from unlabeled video.
The main idea is to use cycle-consistency in time as free supervisory signal for learning visual representations from scratch. 
At training time, our model learns a feature map representation to be useful for performing cycle-consistent tracking. At test time, we use the acquired representation to find nearest neighbors across space and time. We demonstrate the generalizability of the representation -- without finetuning -- across a range of visual correspondence tasks, including video object segmentation, keypoint tracking, and optical flow. Our approach outperforms previous self-supervised methods and performs competitively with strongly supervised methods.
\footnote{Project page: \url{http://ajabri.github.io/timecycle}}

\end{abstract}

\begin{figure*}[t!]
{
\centering
\includegraphics[clip,width=\textwidth]{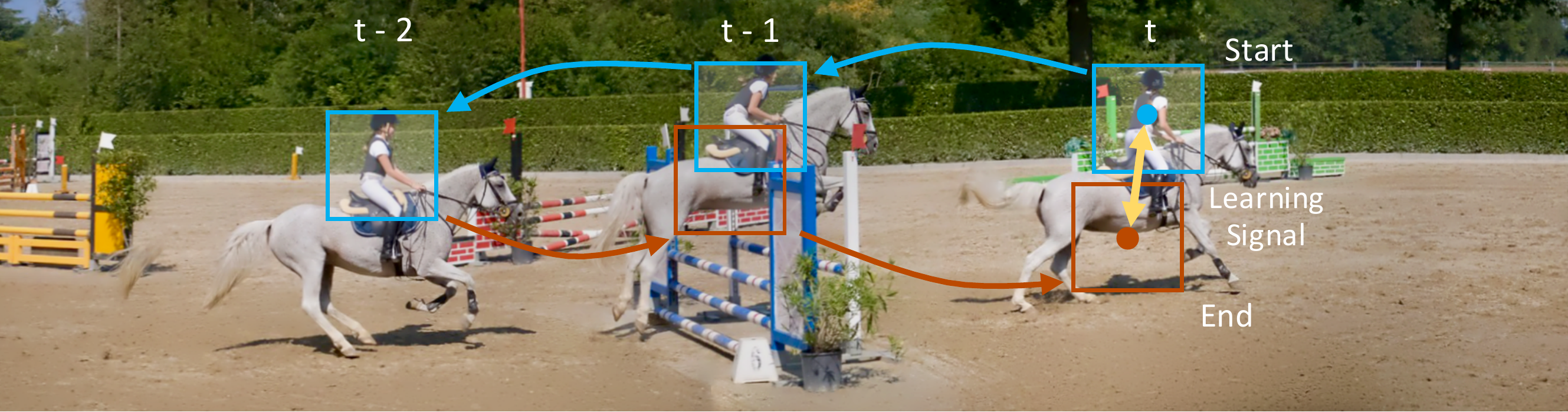}%
 \vspace{-0.05in}
 \caption{\textbf{A Cycle in Time.} Given a video, tracking along the sequence formed by a cycle in time can be self-supervised: the target is simply the beginning  of  the  cycle.  The {yellow} arrow between the start and end represents the differentiable learning signal.}
\vspace{-0.1in}
\label{fig:boys_a}
}
\end{figure*}

\vspace{-0.1in}
\section{Motivation}
\vspace{-0.05in}

It is an oft-told story that when a young graduate student asked Takeo Kanade what are the three most important problems in computer vision, Kanade replied: ``Correspondence, correspondence, correspondence!''
Indeed, most fundamental vision problems, from optical flow and tracking to action recognition and 3D reconstruction, require some notion of visual correspondence.  
Correspondence is the glue that links disparate visual percepts into persistent entities and underlies visual reasoning in space and time.

Learning representations for visual correspondence, from pixel-wise to object-level, has been widely explored, primarily with supervised learning approaches requiring large amounts of labelled data. For learning low-level correspondence, such as optical flow, synthetic computer graphics data is often used as supervision~\cite{fischer2015flownet,IMKDB17,spynet,sun2018pwc}, limiting generalization to real scenes. On the other hand,
approaches for learning higher-level semantic correspondence rely on human annotations~\cite{wang2013learning,held2016learning,valmadre2017end}, which becomes prohibitively expensive at large scale. 
In this work, our aim is to learn representations that support reasoning at various levels of visual correspondence (Figure~\ref{teaser}) from scratch and without human supervision. 



A fertile source of free supervision is video.  Because the world does not change abruptly, 
there is inherent visual correspondence between observations adjacent in time.
The problem is how to find these correspondences and turn them into a learning signal.  
In a largely static world observed by a stationary camera,
such as a webcam trained on the Eiffel Tower, correspondence is straightforward because nothing moves and capturing visual invariance (to weather, lighting) amounts to supervised metric learning. In the dynamic world, however, change in appearance is confounded by movement in space. Finding correspondence becomes more difficult because capturing visual invariance now requires learning to track, but tracking relies on a model of visual invariance. 
This paper proposes to learn to do both simultaneously, in a self-supervised manner.

The key idea is that we can obtain unlimited supervision for correspondence by tracking backward and then forward (i.e. along a cycle in time) and using the {\em inconsistency} between the start and end points as the loss function 
(Figure~\ref{fig:boys_a}). 
We perform tracking by template-matching in a learned deep feature space.
To minimize the loss -- i.e. to be \textit{cycle-consistent} -- the model must learn a feature representation that supports identifying correspondences across frames. As these features improve, the ability to track improves, inching the model toward cycle-consistency. Learning to chain correspondences in such a feature space should thus yield a visual similarity metric tolerant of local transformations in time, which can then be used at test-time as a stand-alone distance metric for correspondence.

\begin{figure*}
{
  \centering
  \includegraphics[width=1.9\columnwidth]{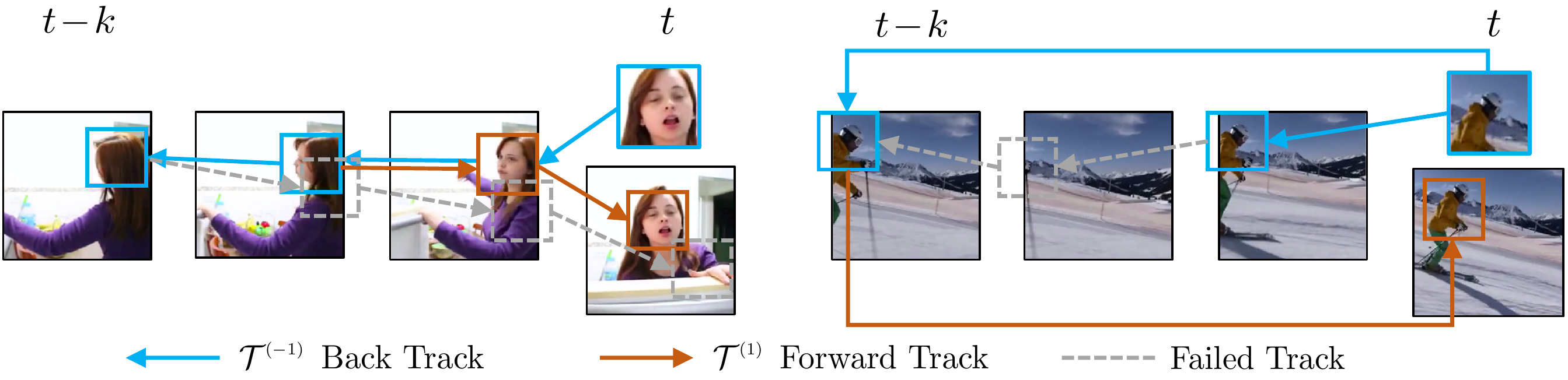}%
  \vspace{-0.08in}
  \caption{{\bf Multiple Cycles and Skip Cycles.}
  Cycle-consistency may not be achievable due to sudden changes in object pose or occlusions. Our solution is to optimize multiple cycles of different lengths simultaneously.  This allows learning from shorter cycles when the full cycle is too difficult (left).  This also allows cycles that skip frames, which can deal with momentary occlusions (right).}
  \label{fig:duality}
  \vspace{-.1in}
}
\end{figure*}

While conceptually simple, implementing objectives based on cycle-consistency can be challenging. 
Without additional constraints, learning can take shortcuts, making correspondences cycle-consistent but wrong~\cite{zhou2016learning}.
In our case, a track that never moves is inherently cycle-consistent. We avoid this by forcing the tracker to re-localize the next patch in each successive frame.
Furthermore, cycle-consistency may not be achievable due to sudden changes in object pose or occlusions; {\textit{skip}}-cycles can allow for cycle-consistency by skipping frames, as in Figure~\ref{fig:duality} (right). 
Finally, correspondence may be poor early in training, and shorter cycles may ease learning, as in Figure~\ref{fig:duality} (left). Thus, we simultaneously learn from many kinds of cycles to induce a natural curriculum and provide better training data.


The proposed formulation can be used with any differentiable tracking operation, providing a general framework for learning representations for visual correspondence from raw video.  Because the method does not rely on human annotation, it can learn from the near infinite video data available online. 
We demonstrate the usefulness of the learned features for tasks at various levels of visual correspondence, ranging from pose, keypoint, and segmentation propagation (of objects and parts) to optical flow.




\section{Related Work}

\paragraph{Temporal Continuity in Visual Learning.}
Temporal structure serves as a useful signal for learning because the visual world is continuous and smoothly-varying. 
Spatio-temporal stability is thought to play a crucial role in the development of invariant representations in biological vision~\cite{yi_spatiotemporal_2008, li_unsupervised_2008, wood_smoothness_2016, wood_development_2016}. For example, Wood~\cite{wood_smoothness_2016} showed that for newborn chicks raised in a visual world that was not temporally smooth, object recognition abilities were severely impaired. 
Computational approaches for unsupervised learning have sought to leverage this continuity, such as continuous transformation learning~\cite{foldiak_learning_1991, wallis_spatio-temporal_1998},  ``slow" feature  learning~\cite{wiskott_slow_2002, zou_deep_2012, Jayaraman2015} and information maximization between neighbouring patches in time~\cite{cpc}. Our work can be seen as slow feature learning with fixation,
learned end-to-end without supervision. 

\vspace{-0.1in}
\paragraph{Self-supervised Representation Learning from Video.} 
Learning representations from video using time as supervision has been extensively studied, both as future prediction task~\cite{Goroshin2015,Srivastava2015LSTMs,Mathieu2015,luo2017unsupervised} as well as motion estimation~\cite{Agrawal2015,Jayaraman2015,tung2017self,Li2016,liu2018switchable}. 
Our approach is most related to the methods of Wang et al.~\cite{Wang_UnsupICCV2015,Wang_UnsupICCV2017} and Pathak et al.~\cite{Pathak2017}, which use off-the-shelf tools for tracking and optical flow respectively, to provide supervisory signal for training. 
However, representations learned in this way are inherently limited by the power of these off-the-shelf tools as well as their failure modes. 
We address this issue by learning the representation and the tracker jointly, and find the two learning problems to be complementary.
Our work is also inspired by the innovative approach of Vondrick et al~\cite{vondrick2018tracking} where video colorization is used as a pretext self-supervised task for learning to track. While the idea is very intriguing, in Section~\ref{sec:experiments} we find that colorization is a weaker source of supervision for correspondence than cycle-consistency, potentially due to the abundance of constant-color regions in natural scenes.  

\vspace{-0.1in}
\paragraph{Tracking.}
Classic approaches to tracking treat it as a matching problem, where the goal is to find a given object/patch in the next frame (see~\cite{FP2012} for overview), and the key challenge is to track reliably over extended time periods~\cite{sethi_finding_1987, wu_situ_2007, multifold_2009, kalal2010forward}. 
Starting with the seminal work of Ramanan et al.~\cite{ramanan2005strike}, researchers largely turned to ``tracking as repeated recognition'', where trained object detectors are applied to each frame independently~\cite{andriluka2008people,kalal2012tracking,wu2013online,wang2013learning,held2016learning,li2018high,valmadre2017end}.
Our work harks back to the classic tracking-by-matching methods in treating it as a correspondence problem, but uses learning to obtain a robust representation that is able to model wide range of appearance changes.

\vspace{-0.1in}
\paragraph{Optical Flow.} Correspondence at the pixel level -- mapping where each pixel goes in the next frame -- is the optical flow estimation problem. Since the energy minimization framework of Horn and Schunck~\cite{horn1981determining} and coarse-to-fine image warping by Lucas and Kanade~\cite{lucas1981iterative},
much progress has been made in optical flow estimation~\cite{memin1998dense,brox2004high,sun2010secrets,fischer2015flownet,IMKDB17,spynet,sun2018pwc}.  However, 
these methods still struggle to scale to long-range correspondence in dynamic scenes with partial observability. These issues have driven researchers to study methods for estimating long-range optical flow~\cite{sand2008particle,brox2009large,rubinstein2012towards,revaud2015epicflow,revaud2016deepmatching,lezama2011track}. For example, Brox and Malik~\cite{brox2009large} introduced a descriptor that matches region hierarchies and provides dense and subpixel-level estimation of flow.  Our work can be viewed as enabling mid-level optical flow estimation.

\vspace{-0.1in}
\paragraph{Mid-level Correspondence.} 

Given our focus on finding correspondence at the patch level, our method is also related to the classic SIFT Flow~\cite{liu2011sift} algorithm and other methods for finding mid-level correspondences between regions across different scenes~\cite{kim2013deformable,ham2016proposal,zhou2015flowweb}. More recently, researchers have studied modeling correspondence in deep feature space~\cite{ufer2017deep,novotny2017anchornet,kim2017fcss,han2017scnet,rocco2017,rocco2018end}. In particular, our work draws from Rocco et al.~\cite{rocco2017,rocco2018end}, who propose a differentiable soft inlier score  for evaluating quality of alignment between spatial features and provides a loss for learning semantic correspondences. Most of these methods rely on learning from simulated or large-scale labeled datasets such as ImageNet, or smaller custom human-annotated data with narrow scope. We address the challenge of learning representations of correspondence without human annotations. 

\vspace{-0.1in}
\paragraph{Forward-Backward and Cycle Consistency.}
Our work is influenced by the classic idea of forward-backward consistency in tracking~\cite{sethi_finding_1987, wu_situ_2007, multifold_2009, kalal2010forward}, which has long been used as an evaluation metric for tracking~\cite{kalal2010forward} as well as a measure of uncertainty~\cite{multifold_2009}. Recent work on optical flow estimation~\cite{mahajan2009moving,janai2018unsupervised,vijayanarasimhan2017sfm,wang2018occlusion,meister2017unflow}  also utilizes forward-backward consistency as an optimization goal. For example, Meister et al.~\cite{meister2017unflow} combines one-step forward and backward consistency check with pixel reconstruction loss for learning optical flows. Compared to pixel reconstruction, modeling correspondence in feature space allows us to follow and learn from longer cycles. 
Forward-backward consistency is a specific case of cycle-consistency, which has been widely applied as a learning objective for 3D shape matching~\cite{huang2013consistent}, image alignment~\cite{zhou2015flowweb,zhou2015multi,zhou2016learning}, depth estimation~\cite{zhou2017unsupervised,godard2017unsupervised,yin2018geonet},
and image-to-image translation~\cite{zhu2017unpaired,Recycle-GAN}. For example Zhou et al.~\cite{zhou2016learning} used 3D CAD models to render two synthetic views for pairs of training images and construct a correspondence flow 4-cycle. To the best of our knowledge, our work is the first to employ cycle-consistency across multiple steps in time. 

 \section{Approach}

\begin{figure*}
{
    \centering
    \begin{subfigure}[b]{0.4\textwidth}
        \centering
        \includegraphics[width=1\columnwidth]{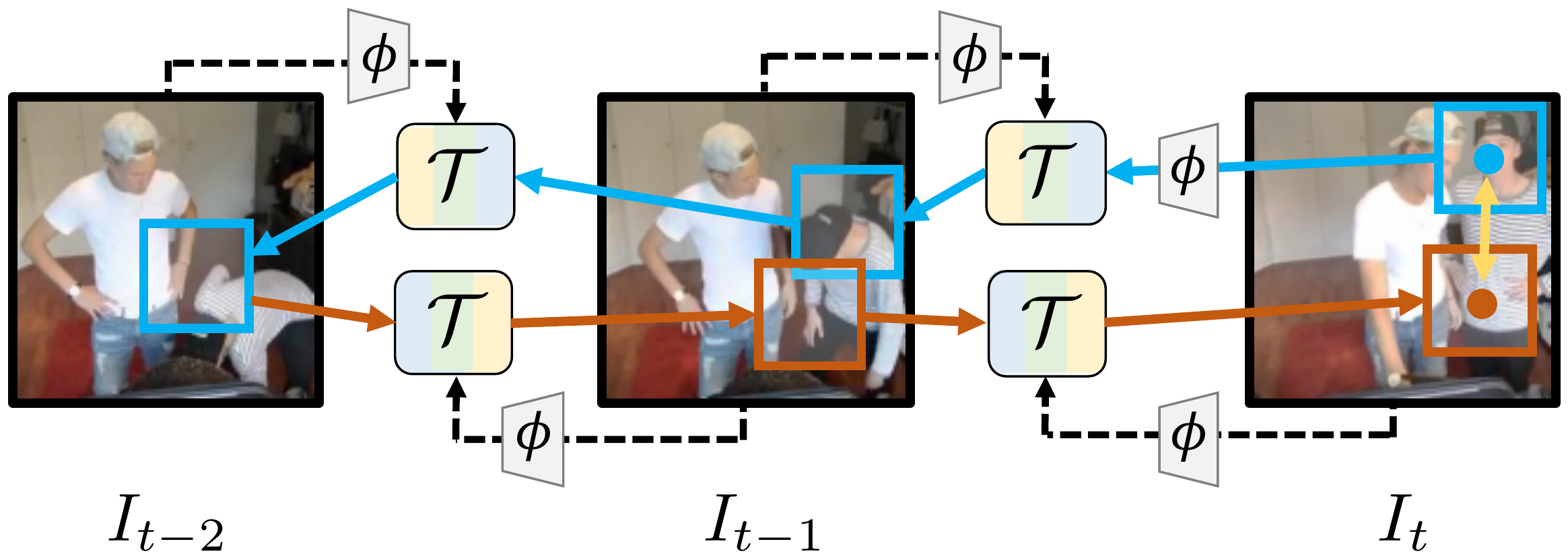}
        \vspace{2mm}
        \caption{Training $\phi$ by End-to-end Cycle-consistent Tracking}
      \label{fig:overview}
    \end{subfigure}%
    ~ 
    \begin{subfigure}[b]{0.6\textwidth}
        \centering
        \includegraphics[width=1\columnwidth]{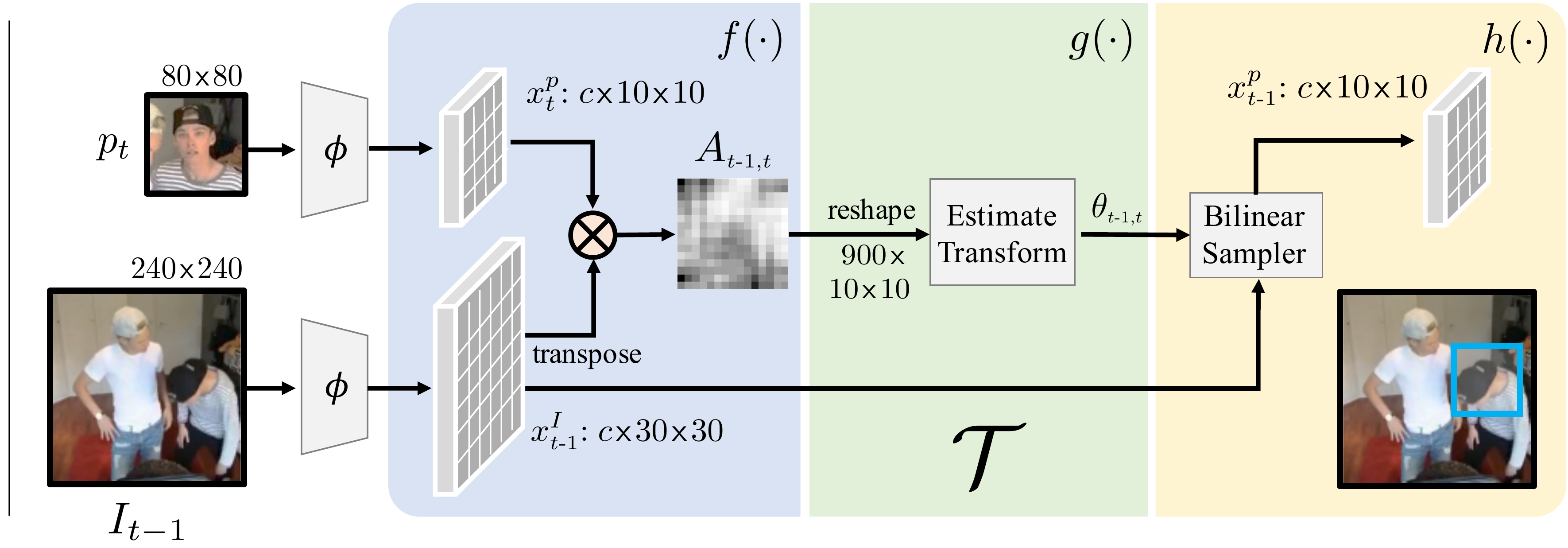}
        \caption{Differentiable Tracking Operation $\mathcal{T}$}
      \label{fig:tracker}

    \end{subfigure}
\vspace{-.25in}
    \caption{\textbf{Method Overview}. {(a) During training, the model learns a feature space encoded by $\phi$ to perform tracking using tracker $\mathcal{T}$. By tracking backward and then forward, we can use cycle-consistency to supervise learning of $\phi$.} Note that only the initial patch $p_t$ is explicitly encoded by $\phi$; other patch features along the cycle are obtained by localizing image features. (b) We show one step of tracking back in time from $t$ to $t-1$. Given input image features $x^I_{t-1}$ and query patch features $x^p_t$, $\mathcal{T}$ localizes the patch $x^p_{t-1}$ in $x^I_{t-1}$. This operation is performed iteratively to track along the cycle in (a).}
    \label{fig:method}
    \vspace{-4mm}
}
\end{figure*}

An overview of the training procedure is presented in Figure~\ref{fig:overview}. The goal is to learn a feature space $\phi$ by tracking a patch $p_t$ extracted from image $I_t$ backwards and then forwards in time, while minimizing the cycle-consistency loss $l_\theta$ (yellow arrow). 
Learning $\phi$ relies on a simple tracking operation $\mathcal{T}$, which takes as inputs the features of a current patch and a target image, and returns the image feature region with maximum similarity. Our implementation of $\mathcal{T}$ is shown in Figure~\ref{fig:tracker}: without information of where the patch came from, $\mathcal{T}$ must match features encoded by $\phi$ to localize the next patch. As shown in Figure~\ref{fig:overview}, $\mathcal{T}$ can be iteratively applied backwards and then forwards through time to track along an arbitrarily long cycle. The  cycle-consistency loss $l_\theta$ is the euclidean distance between the spatial coordinates of initial patch $p_t$ and the patch found at the end of the cycle in $I_t$. In order to minimize $l_\theta$, the model must learn a feature space $\phi$ that allows for robustly measuring visual similarity between patches along the cycle. 

Note that $\mathcal{T}$ is \textit{only used in training} and is deliberately designed to be weak, so as to place the burden of representation on $\phi$. At test time, the learned $\phi$ is used directly for computing correspondences.
In the following, we first formalize cycle-consistent tracking loss functions and then describe our architecture for mid-level correspondence.

\subsection{Cycle-Consistency Losses}
\vspace{-0.05in}
We describe a formulation of cycle-consistent tracking and use it to succinctly express loss functions based on temporal cycle-consistency.

\vspace{-0.1in}
\subsubsection{Recurrent Tracking Formulation}
\vspace{-0.05in}
Consider as inputs a sequence of video frames $I_{t-k:t}$ and a patch $p_t$ taken from $I_t$. These pixel inputs are mapped to a feature space by an encoder $\phi$, such that $x^I_{t-k:t} = \phi(I_{t-k:t})$ and $x^p_t = \phi(p_t)$.

Let $\mathcal{T}$ be a differentiable  operation $ x^I_s \times x^p_t \mapsto {x}^p_s$, where $s$ and $t$ represent time steps. The role of $\mathcal{T}$ is to localize the patch features $x^p_s$ in image features $x^I_s$ that are most similar to $x^p_t$. 
We can apply  $\mathcal{T}$ iteratively in a forward manner $i$ times from $t-i$ to $t-1$: 
$$\mathcal{T}^{(i)}(x^I_{t-i}, x^p) = \mathcal{T}(x^I_{t-1}, \mathcal{T}(x^I_{t-2}, ... \mathcal{T}(x^I_{t-i}, x^p)))$$
By convention, the tracker $\mathcal{T}$ can be applied backwards $i$ times from time $t-1$ to $t-i$:
$$\mathcal{T}^{(-i)}(x^I_{t-1}, x^p) = \mathcal{T}(x^I_{t-i}, \mathcal{T}(x^I_{t-i+1}, ... \mathcal{T}(x^I_{t-1}, x^p)))$$

\subsubsection{Learning Objectives}
The following learning objectives rely on a measure of agreement  $l_\theta(x^p_t, \hat{x}^p_t)$ between the initial patch and re-localized patch  (defined in Section~\ref{ssec:specs}). 
\vspace{1mm}

\textbf{Tracking}: The cycle-consistent loss $\mathcal{L}^i_{long}$ is defined as
$$\mathcal{L}^i_{long} = l_\theta(x^p_t, \mathcal{T}^{(i)}(x^I_{t-i+1}, \mathcal{T}^{(-i)}(x^I_{t-1}, x^p_t))).$$
The tracker attempts to follow features backward and then forward $i$ steps in time to re-arrive to the initial query, as depicted in Figure \ref{fig:overview}.
\vspace{1mm}

\textbf{Skip Cycle:}  In addition to cycles through consecutive frames, we also allow skipping through time. We define the loss on a two-step skip-cycle as $\mathcal{L}^i_{skip}$:
\begin{align*}
\mathcal{L}^i_{skip} = l_\theta(x^p_t, \mathcal{T}(x^I_{t}, \mathcal{T}(x^I_{t-i}, x^p_t))).
\end{align*}
This attempts longer-range matching by skipping to the frame $i$ steps away. 
\vspace{1mm}

\textbf{Feature Similarity: } We explicitly require the query patch $x^p_t$ and localized patch $\mathcal{T}(x^I_{t-i}, x^p_t)$ to be similar in feature space. This loss amounts to the negative Frobenius inner product between spatial feature tensors: 
\begin{align*}
\vspace{-2mm}
\mathcal{L}^i_{sim} &= - \langle{x^p_t} , {\mathcal{T}(x^I_{t-i}, x^p_t)}\rangle 
\end{align*}
In principle, this loss can further be formulated as the inlier loss from~\cite{rocco2018end}.
The overall learning objective sums over the $k$ possible cycles, with weight $\lambda=0.1$: 
\begin{align*}
    \mathcal{L} &= \sum_{i=1}^k \mathcal{L}^i_{sim} + \lambda \mathcal{L}^i_{skip} + \lambda\mathcal{L}^i_{long}.
\end{align*}

\vspace{-0.05in}
\subsection{Architecture for Mid-level Correspondence }\label{ssec:specs}
The learning objective thus described can be used to train arbitrary differentiable tracking models. In practice, the architecture of the encoder determines the type of correspondence captured by the acquired representation. 
In this work, we are interested in a model for mid-level temporal correspondence. Accordingly, we choose the representation to be a mid-level deep feature map, coarser than pixel space but with sufficient spatial resolution to support tasks that require localization. An overview is provided in Figure \ref{fig:tracker}. 

\vspace{-0.05in}
\subsubsection{Spatial Feature Encoder $\phi$}
\vspace{-0.05in}
We compute spatial features with a ResNet-50 architecture~\cite{He2016} without res$_5$ (the final 3 residual blocks). We reduce the spatial stride of res$_4$ 
for larger spatial outputs. Input frames are $240\times240$ pixels, randomly cropped from video frames re-scaled to have $\texttt{min}(H, W) = 256$. The size of the spatial feature of the frame is thus $30 \times 30$. Image patches are $80\times80$, randomly cropped from the full $240\times240$ frame, so that the feature is $10 \times 10$. We perform $l_2$ normalization on the channel dimension of spatial features to facilitate computing cosine similarity.

\vspace{-0.05in}
\subsubsection{Differentiable Tracker $\mathcal{T}$}
\vspace{-0.05in}
Given the representation from the encoder, we perform tracking with $\mathcal{T}$. As illustrated in Figure~\ref{fig:tracker}, the differentiable tracker is composed of three main components. 

\vspace{-0.05in}
\paragraph{Affinity function $f$} provides a measure of similarity between coordinates of spatial features $x^I$ and $x^p$. We denote the affinity function as $f(x^I, x^p) := A$, such that  
$f : \mathbb{R}^{c\times 30 \times 30} \times \mathbb{R}^{c\times 10 \times 10}  \xrightarrow{} \mathbb{R}^{ 900 \times  100}$.

A generic choice for computing the affinity is the dot product between embeddings, referred to in recent literature as attention~\cite{Vaswani2017,Wang_nonlocalCVPR2018} and more historically known as normalized cross-correlation~\cite{fischer2015flownet,li2018high}. With spatial grid $j$ in feature $x^I$ as $x^I(j)$ and the grid $i$ in $x^p$ as $x^p(i)$,
\begin{equation}\label{eq:sim}
A(j,i) = \frac{\exp{(x^I(j)^\intercal x^p(i))}}{\sum_j \exp{(x^I(j)^\intercal x^p(i))}}
\end{equation}
where the similarity $A(j,i)$ is normalized by the softmax over the spatial dimension of $x^I$, for each $x^p(i)$. Note that the affinity function is defined for any feature dimension. 

\vspace{-0.05in}
\paragraph{Localizer $g$} takes affinity matrix $A$ as input and estimates localization parameters $\theta$ corresponding to the patch in feature $x^I$ which best matches $x^p$. $g$ is composed of two convolutional layers and one linear layer. We restrict $g$ to output 3 parameters for the bilinear sampling grid (i.e. simpler than~\cite{stn}), corresponding to 2D translation and rotation: $g(A) := \theta$, where $g : \mathbb{R}^{900 \times 100} \xrightarrow{} \mathbb{R}^3$. The expressiveness of $g$ is intentionally limited so as to place the burden of representation on the encoder (see Appendix~\ref{appendix:b}).

\vspace{-0.05in}
\paragraph{Bilinear Sampler $h$} uses the image feature $x^I$ and $\theta$ predicted by $g$ to perform bilinear sampling to produce a new patch feature $h(x^I, \theta)$ which is in the same size as $x^p$, such that $h : \mathbb{R}^{c\times 30 \times 30} \times \mathbb{R}^{3}  \xrightarrow{} \mathbb{R}^{c\times 10 \times 10}$. 

\vspace{-0.1in}
\subsubsection{End-to-end Joint Training} 
\vspace{-0.05in}
The composition of encoder $\phi$ and $\mathcal{T}$ forms a differentiable patch tracker, allowing for end-to-end  training of $\phi$ and $\mathcal{T}$:
\vspace{-0.05in}
\begin{align*}
x^I, x^p &= \phi(I), \phi(p) \\
\mathcal{T}(x^I, x^p) &= h(x^I, g(f(x^I, x^p)).
\end{align*}

\vspace{-0.2in}
\paragraph{Alignment Objective $l_\theta$} is applied in the cycle-consistent losses $\mathcal{L}^i_{long}$ and $\mathcal{L}^i_{skip}$, measuring the error in alignment between two patches. We follow the formulation introduced by \cite{rocco2017}. Let $M(\theta_{x^p})$ correspond to the bilinear sampling grids used to form a patch feature $x^p$ from image feature $x^I$. Assuming $M(\theta_{x^p})$ contains $n$ sampling coordinates, the alignment objective is defined as:
\vspace{-0.05in}
\begin{align*}
    l_\theta(x^p_*, \hat{x}^p_t) = \frac{1}{n} \sum_{i=1}^{n}   ||M(\theta_{x^p_*})_i - M(\theta_{\hat{x}^p_t})_i||^2_2
\end{align*}

\vspace{-0.1in}
\section{Experiments}
\label{sec:experiments}
\vspace{-0.05in}
\begin{figure*}
{
  \vspace{-2mm}
  \centering
  \includegraphics[width=2.0\columnwidth]{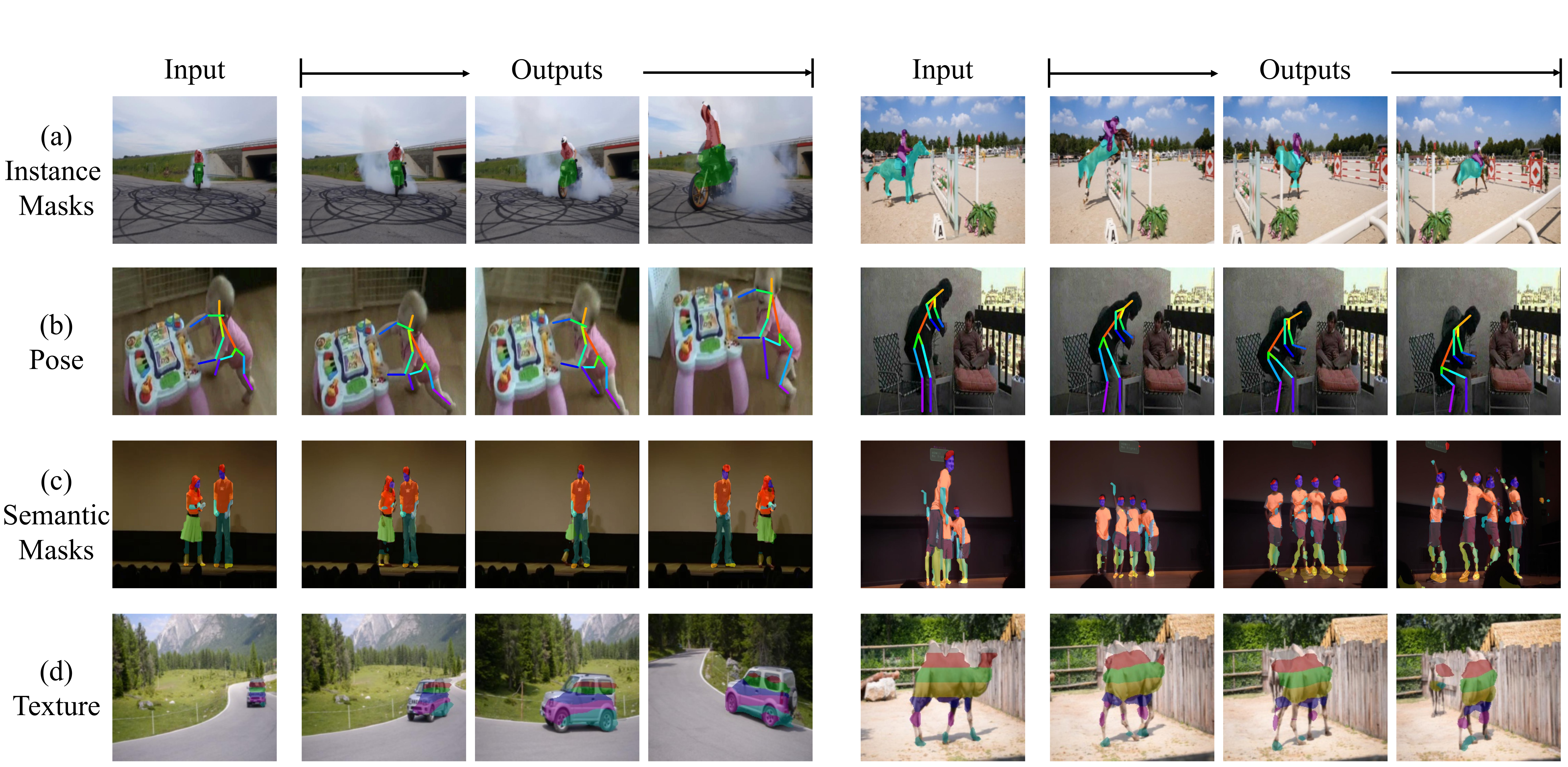}%
  \vspace{-0.1in}
  \caption{Visualizations of our propagation results. Given the labels as input in the first frame, our feature can propagate them to the rest of frames, without further fine-tuning. The labels include (a) instance masks in DAVIS-2017~\cite{Pont-Tuset_arXiv_2017}, (b) pose keypoints in JHMDB~\cite{Jhuang:ICCV:2013}, (c) semantic masks in VIP~\cite{zhouACMMM2018} and even (d) texture map. }
  \label{fig:result}
  \vspace{-0.15in}  
}
\end{figure*}

We report experimental results for a model trained on the VLOG dataset~\cite{Fouhey18} from scratch; training on other large video datasets such as Kinetics gives similar results (see Appendix~\ref{appendix:kinetics}). The trained representation is evaluated \textit{without fine-tuning} on several challenging video propagation tasks: DAVIS-2017~\cite{Pont-Tuset_arXiv_2017}, JHMDB~\cite{Jhuang:ICCV:2013} and Video Instance-level Parsing (VIP)~\cite{zhouACMMM2018}. Through various experiments, we show that the acquired representation generalizes to a range of visual correspondence tasks (see Figure~\ref{fig:result}). 

\subsection{Common Setup and Baselines}
\vspace{-0.05in}
\textbf{Training.} We train the model on the VLOG dataset~\cite{Fouhey18} without using any annotations or pre-training. The VLOG dataset contains 114K videos and the total length of the videos is 344 hours. During training, we set the number of past frames as $k=4$. We train on a 4-GPU machine with a mini-batch size of 32 clips (8 clips per GPU), for 30 epochs. The model is optimized with Adam~\cite{kingma2014adam} with a learning rate of $0.0002$ and momentum term $\beta_1 = 0.5, \beta_2 = 0.999$.

\textbf{Inference.} At test time, we use the trained encoder's representation to compute dense correspondences for video propagation. Given initial labels of the first frame, we propagate the labels to the rest of the frames in the video. Labels are given by specified targets for the first frame of each task, with instance segmentation masks for DAVIS-2017~\cite{Pont-Tuset_arXiv_2017}, human pose keypoints JHMDB~\cite{Jhuang:ICCV:2013}, and both instance-level and semantic-level masks for VIP~\cite{zhouACMMM2018}. The labels of each pixel are discretized to $C$ classes. For segmentation masks, $C$ is the number of instance or semantic labels. For keypoints, $C$ is the number of keypoints. We include a background class. We propagate the labels in the feature space. The labels in the first frame are one-hot vectors, while propagated labels are soft distributions. 

\textbf{Propagation by k-NN.} Given a frame $I_t$ and a frame $I_{t-1}$ with labels, we compute their affinity in feature space: $A_{t-1,t} = f(\phi(I_{t-1}), \phi(I_{t}))$ (Eq.~\ref{eq:sim}). We compute label $y_i$ of pixel $i$ in  $I_t$ as
\vspace{-3mm}
\begin{align}
\label{eq:knn}
\small{
y_i = \sum_{j} A_{t-1,t}(j, i) y_j,
}
\end{align}

\vspace{-3mm}
\hspace{-4mm}where $A_{t-1,t}(j, i)$ is the affinity between pixels $i$ in $I_t$ and  $j$ in $I_{t-1}$. We propagate from the top-$5$ pixels with the greatest affinity $A_{t-1,t}(j, i)$ for each pixel $i$. Labels are propagated from $I_{t-1:t-K}$, as well as $I_1$, and averaged.  Finally, we up-sample the label maps to image size. For segmentation, we use the \texttt{argmax} of the class distribution of each pixel. For keypoints, we choose the pixel with the maximum score for each keypoint type. 

\vspace{-4mm}
\paragraph{Baselines.} We compare with the following baselines: 
\begin{itemize}
\vspace{-0.05in}
\setlength\parskip{-0.1em}
    \item \textbf{Identity}: Always copy the first frame labels.
    \item \textbf{Optical Flow} (FlowNet2~\cite{IMKDB17}): A state-of-the-art method for predicting optical flow with neural networks ~\cite{IMKDB17}. We adopt the open-source implementation which is trained with synthetic data in a supervised manner. For a target frame $I_t$, we compute the optical flow from frame $I_{t-1}$ to $I_t$ and warp the labels in $I_{t-1}$ to $I_t$. 
    \item \textbf{SIFT Flow}~\cite{liu2011sift}: For a target frame $I_t$, we compute the SIFT Flow between $I_t$ and its previous frames. We propagate the labels in $K$ frames before $I_t$ and the first frame via SIFT Flow warping. The propagation results are averaged to compute the labels for $I_t$.
    \item \textbf{Transitive Invariance}~\cite{Wang_UnsupICCV2017}: A self-supervised approach that combines multiple objectives: (i) visual tracking on raw video~\cite{Wang_UnsupICCV2015} and (ii) spatial context reasoning~\cite{doersch2015unsupervised}. We use the open-sourced pre-trained VGG-16~\cite{simonyan2014very} model and adopt our proposed inference procedure.
    \item \textbf{DeepCluster}~\cite{caron2018deep}: A self-supervised  approach which uses a K-means objective to iteratively update targets and learn a mapping from images to targets. It is trained on the ImageNet dataset without using annotations. We apply the trained model with VGG-16 and adopt the same inference procedure as our method.
    \item \textbf{Video Colorization}~\cite{vondrick2018tracking}: A self-supervised  approach for label propagation. Trained on the Kinetics~\cite{Kay2017} dataset, it uses color propagation as self-supervision. The architecture is based on 3D ResNet-18. We report their results.
    \item \textbf{ImageNet Pre-training}~\cite{He2016}: The conventional setup for supervised training of ResNet-50 on ImageNet.
    \item \textbf{Fully-Supervised Methods}: We report fully-supervised methods for reference, which not only use ImageNet pre-training but also fine-tuning on the target dataset. Note that these methods do not always follow the inference procedure used with method, and labels of the first frame are not used for JHMDB and VIP at test time.
\end{itemize}

\subsection{Instance Propagation on DAVIS-2017}

\begin{table}[t]
\centering
\small
\tablestyle{6pt}{1.05}
\begin{tabular}{l|c|x{24}x{24}}
\multicolumn{1}{c|}{model}  & Supervised  & $\mathcal{J}$(Mean) & $\mathcal{F}$(Mean) \\
\shline
Identity &  & 22.1 & 23.6 \\
Random Weights (ResNet-50) &  & 12.4 & 12.5 \\
Optical Flow (FlowNet2)~\cite{IMKDB17} &  & 26.7 & 25.2 \\
SIFT Flow~\cite{liu2011sift} &  & 33.0 & 35.0 \\
Transitive Inv.~\cite{Wang_UnsupICCV2017} &  & 32.0 & 26.8 \\
DeepCluster~\cite{caron2018deep} &  & 37.5 & 33.2 \\
Video Colorization~\cite{vondrick2018tracking} &  & 34.6 & 32.7 \\
Ours (ResNet-18)&  & 40.1 & 38.3 \\
Ours (ResNet-50)&  & \textbf{41.9} & \textbf{39.4} \\
\hline
ImageNet (ResNet-50)~\cite{He2016}   & \checkmark & 50.3 &  49.0  \\
Fully Supervised~\cite{yang2018efficient,Cae17}   & \checkmark & 55.1 &  62.1  \\
\end{tabular}
\vspace{-0.05in}
\caption{Evaluation on instance mask propagation on  DAVIS-2017~\cite{Pont-Tuset_arXiv_2017}. We follow the standard metric on region similarity $\mathcal{J}$ and contour-based accuracy $\mathcal{F}$. }
\label{tab:davis}
\vspace{-0.10in}
\end{table}

We apply our model to video object segmentation on the DAVIS-2017 validation set~\cite{Pont-Tuset_arXiv_2017}. Given the initial masks of the first frame, we propagate the masks to the rest of the frames. Note that there can be multiple instances in the first frame. We follow the standard metrics including the region similarity $\mathcal{J}$ (IoU) and the contour-based accuracy $\mathcal{F}$.  We set $K=7$, the number of reference frames in the past.

We show comparisons in Table~\ref{tab:davis}. Comparing to the recent Video Colorization approach~\cite{vondrick2018tracking}, our method is $7.3\%$ in $\mathcal{J}$ and $6.7\%$ in $\mathcal{F}$. Note that although we are only $4.4\%$ better than the DeepCluster baseline in $\mathcal{J}$, we are better in contour accuracy $\mathcal{F}$ by $6.2\%$. Thus, DeepCluster does not capture dense correspondence on the boundary as well. 

For fair comparisons, we also implemented our method with a ResNet-18 encoder, which has less parameters compared to the VGG-16 in~\cite{Wang_UnsupICCV2017,caron2018deep} and the 3D convolutional ResNet-18 in~\cite{vondrick2018tracking}. We observe that results are only around $2\%$ worse than our model with ResNet-50, which is still better than the baselines. 

While the ImageNet pre-trained network performs better than our method on this task, we argue it is easy for the ImageNet pre-trained network to recognize objects under large variation as it benefits from curated object-centric annotation. Though our model is only trained on indoor scenes without labels, it generalizes to outdoor scenes.

Although video segmentation is an important application, it does not necessarily show that the representation captures dense correspondence.

\subsection{Pose Keypoint Propagation on JHMDB}

To see whether our method is learning more spatially precise correspondence, we apply our model on the task of keypoint propagation on the split 1 validation set of JHMDB~\cite{Jhuang:ICCV:2013}. Given the first frame with $15$ labeled human keypoints, we propagate them through time. We follow the evaluation of the standard PCK metric~\cite{yang2013articulated}, which measures the percentage of keypoints close to the ground truth in different thresholds of distance. We set the number of reference frames same as experiments in DAVIS-2017.

As shown in Table~\ref{tab:jhmdb}, our method outperforms all self-supervised baselines by a large margin. We observe that SIFT Flow actually performs better than other self-supervised learning methods in PCK@.1. Our method outperforms SIFT Flow by $8.7\%$ in PCK@.1 and $9.9\%$ in PCK@.2. Notably, our approach is only $0.7\%$ worse than ImageNet pre-trained features in PCK@.1 and performs better in PCK@.2.

\begin{table}[t]
\centering
\small
\tablestyle{6pt}{1.05}
\begin{tabular}{l|c|x{24}x{24}}
\multicolumn{1}{c|}{model}  & Supervised  & PCK@.1 & PCK@.2 \\
\shline
Identity &   &  43.1 & 64.5 \\
Optical Flow (FlowNet2)~\cite{IMKDB17} &  &   45.2 & 62.9 \\
SIFT Flow~\cite{liu2011sift} &   &  49.0 & 68.6 \\
Transitive Inv.~\cite{Wang_UnsupICCV2017} &   & 43.9 & 67.0 \\
DeepCluster~\cite{caron2018deep} &   & 43.2 & 66.9 \\
Video Colorization~\cite{vondrick2018tracking} &   & 45.2 & 69.6 \\
Ours (ResNet-18)&   & 57.3 & 78.1 \\
Ours (ResNet-50)&   & \textbf{57.7} & \textbf{78.5} \\
\hline
ImageNet (ResNet-50)~\cite{He2016}  & \checkmark  & 58.4 &  78.4  \\
Fully Supervised~\cite{song2017thin} & \checkmark   & 68.7 &  92.1  \\
\end{tabular}
\vspace{-0.05in}
\caption{Evaluation on pose propagation on JHMDB~\cite{Jhuang:ICCV:2013}. We report the PCK in different thresholds.}
\vspace{-0.14in}
\label{tab:jhmdb}
\end{table}

\subsection{Semantic and Instance Propagation on VIP}

We apply our approach on the Video Instance-level Parsing (VIP) dataset~\cite{zhouACMMM2018}, which is densely labeled with semantic masks for different human parts (e.g., hair, right arm, left arm, coat). It also has instance labels that differentiate humans. Most interestingly, the duration of a video ranges from 10 seconds to 120 seconds in the dataset, which is much longer than aforementioned datasets.

We test our method on the validation set of two tasks in this dataset: (i) The first task is to propagate the semantic human part labels from the first frame to the rest of the video, and evaluate with the mean IoU metric; (ii) In the second task, the labels in the first frame are given with not only the semantic labels but also the instance identity. Thus, the model must differentiate the different arms of different human instances. We use the standard instance-level human parsing metric~\cite{li2017holistic}, mean Average Precision, for overlap thresholds varying from 0.1 to 0.9. Since part segments are relatively small (compared to objects in DAVIS-2017), we increase the input image size to $560 \times 560$ for inference, and use two reference frames, including the first frame.

\textbf{Semantic Propagation.} As shown with the mIoU metric in Table~\ref{tab:vip}, our method again exceeds all self-supervised baselines by a large margin (a ~\cite{vondrick2018tracking} model is currently not available). ImageNet pre-trained models have the advantage of semantic annotation and thus do not necessarily have to perform tracking. As shown in Figure~\ref{fig:result}(c), our method is able to handle occlusions and multiple instances. 

\textbf{Part Instance Propagation.} This task is more challenging. We show the results in mean $AP^r_\text{vol}$ in Table~\ref{tab:vip}. Our method performs close to the level of ImageNet pre-trained features. We show different radial thresholds for average precision ($AP^r_\text{vol}$) in Table~\ref{tab:vipablate}. ImageNet pre-trained features performs better under smaller thresholds and worse under larger thresholds, suggesting that it has an advantage in finding coarse correspondence while our method is more capable of spatial precision.

\begin{table}[t]
\centering
\small
\tablestyle{6pt}{1.05}
\begin{tabular}{l|c|x{24}x{24}}
\multicolumn{1}{c|}{model}  & Supervised  & mIoU & $AP^r_\text{vol}$ \\
\shline
Identity &  & 13.6 & 4.0 \\
Optical Flow (FlowNet2)~\cite{IMKDB17} &  & 16.1 & 8.3 \\
SIFT Flow~\cite{liu2011sift} &  & 21.3 & 10.5 \\
Transitive Inv.~\cite{Wang_UnsupICCV2017} &  & 19.4 & 5.0 \\
DeepCluster~\cite{caron2018deep} &  & 21.8 & 8.1 \\
Ours (ResNet-50)&  & \textbf{28.9} & \textbf{15.6} \\
\hline
ImageNet (ResNet-50)~\cite{He2016}   & \checkmark & 34.7 &  16.1  \\
Fully Supervised~\cite{zhouACMMM2018}   & \checkmark & 37.9 &  24.1  \\
\end{tabular}
\vspace{-0.05in}
\caption{Evaluation on propagating human part labels in Video Instance-level Parsing (VIP) dataset~\cite{zhouACMMM2018}. We measure \emph{Semantic Propagation} with mIoU and \emph{Part Instance Propagation} in $AP^r_\text{vol}$.}
\label{tab:vip}
\end{table}

\begin{table}[t]
\vspace{-2mm}
\centering
\small
\tablestyle{6pt}{1.05}
\begin{tabular}{l|c|x{24}x{11}x{15}x{13}}
 & \multirow{2}{*}{$AP^r_\text{vol}$} &\multicolumn{3}{c}{IoU threshold} \\
\multicolumn{1}{c|}{model}  & & 0.3 & 0.5 & 0.7 \\
\shline
Ours (ResNet-50) & 15.6 & 23.0 & \textbf{12.7} & \textbf{5.4} \\
ImageNet (ResNet-50)~\cite{He2016}  & \textbf{16.1} & \textbf{24.2} & 11.9 & 4.8 \\
\end{tabular}
\vspace{-0.1in}
\caption{A more detailed analysis of different thresholds for \emph{Part Instance Propagation} on the VIP dataset~\cite{zhouACMMM2018}.}
\vspace{-0.08in}
\label{tab:vipablate}
\end{table}

\subsection{Texture Propagation}
The acquired representation allows for propagation of not only instance and semantic labels, but also textures. We visualize texture propagation in Figure~\ref{fig:result} (d); these videos are samples from DAVIS-2017~\cite{Pont-Tuset_arXiv_2017}. We ``paint" a texture of 6 colored stripes on an the object in the first frame and propagate it to the rest of the frames using our representation. We observe that the structure of the texture is well preserved in the following frames, demonstrating that the representation allows for finding precise correspondence smoothly though time. See the project page for video examples.

\subsection{Video Frame Reconstructions}

Though we do not optimize for pixel-level objectives at training time, we can evaluate how well our method performs on pixel-level reconstruction. Specifically, given two images $I_s$ and $I_t$ distant in time in a video, we compute coordinate-wise correspondences under the acquired representation and generate a flow field for pixel movement  between $I_s$ and $I_t$. We then upsample the flow field to the same size as the image and warp it on Image $I_s$ to generate a new image $I_t^\prime$ (as shown in Figure~\ref{fig:flowResult}). We compare the L1 distance between $I_t^\prime$ and $I_t$ in RGB space and report the reconstruction errors in Table~\ref{tab:flow}.

\begin{figure}
{
  \centering
  \includegraphics[width=1\columnwidth]{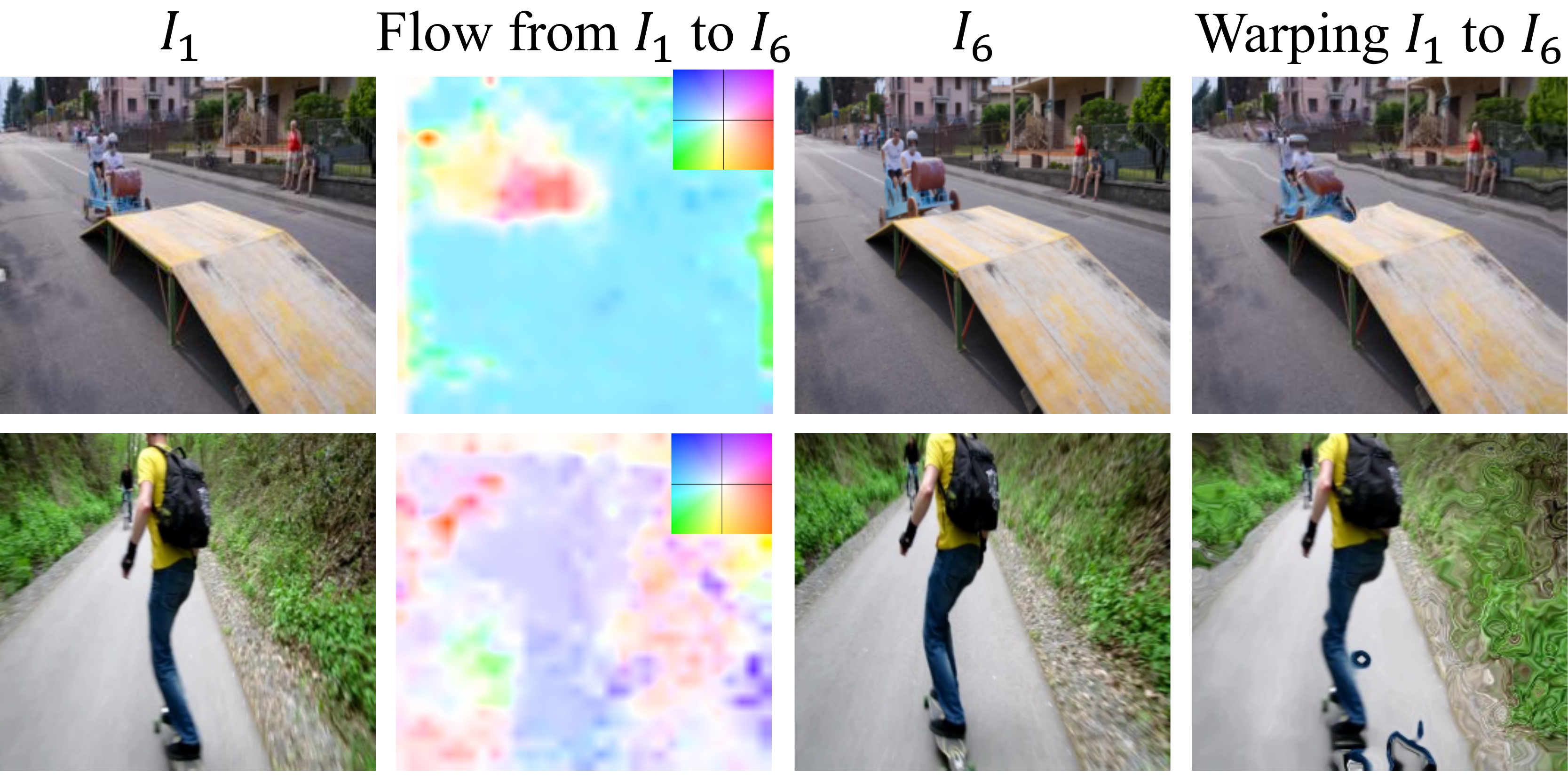}%
  \vspace{-0.02in}
  \caption{Given $I_1, I_6$ which have 5-frame gap, we compute the long-range flows between them with our representation. This flow can be used to warp $I_1$ to generate image similar to $I_6$. }
  \vspace{-0.08in}
  \label{fig:flowResult}
}
\end{figure}

For fair comparison, we perform this experiment on the DAVIS-2017 validation set, which none of the reported methods have seen. We experiment with two time gaps, 5 and 10 frames. For the smaller gap, FlowNet2~\cite{IMKDB17} performs reasonably well, whereas reconstruction degrades for larger gaps. In both cases, our method performs better than FlowNet2 and the ImageNet pre-trained network. This is encouraging: our method is not trained with pixel-level losses, yet out-performs methods trained with pixel-level tasks and human supervision.

\begin{table}[t]
\centering
\small
\tablestyle{10pt}{1.05}
\begin{tabular}{l|c|x{24}x{24}}
\multicolumn{1}{c|}{model}  & 5-F & 10-F \\
\shline
Identity &  82.0 & 97.7 \\
Optical Flow (FlowNet2)~\cite{IMKDB17} &  62.4 & 90.3 \\
ImageNet (ResNet-50)~\cite{He2016}   & 64.0 &  79.2  \\
Ours (ResNet-50)& \textbf{60.4} & \textbf{76.4} \\
\end{tabular}
\vspace{-0.08in}
\caption{We compute the long-range flow on two frames and warp the first one with the flow. We compare the warped frame with the second frame in L1 distance. The gaps are 5 or 10 frames.}
\vspace{-0.1in}
\label{tab:flow}
\end{table}

\vspace{-1mm}
\section{Limitations and Future Work}
While in principle our method should keep improving with more data, in practice, learning seems to plateau after a moderate amount of training (i.e. 30 epochs).    
An important next step is thus how to better scale to larger, noisier data. A crucial component is improving robustness to occlusions and partial observability, for instance, by using a better search strategy for finding cycles at training time. Another issue is deciding \textit{what} to track at training time. Picking patches at random can result in issues such as stationary background patches and tracking ambiguity -- e.g. how should one track a patch containing two objects that eventually diverge? Jointly learning what to track may also give rise to unsupervised object detection.
Finally, incorporating more context for tracking both at training and test time may be important for learning more expressive models of spatial-temporal correspondence.

We hope this work is a step toward learning from the abundance of visual correspondence inherent in raw video in a scalable and end-to-end manner. While our experiments show promising results at certain levels of correspondence, much work remains to cover the full spectrum.

\vspace{-3mm}
{
\paragraph{Acknowledgements:}
We thank members of the BAIR community for helpful discussions and feedback, and Sasha Sax and Michael Janner for draft comments. AJ is supported by the P.D. Soros Fellowship. XW is supported by the Facebook PhD Fellowship. This work was also supported, in part, by NSF grant IIS-1633310 and Berkeley DeepDrive.
}

{\small
\bibliographystyle{ieee}
\bibliography{timecycle}
}

\clearpage

\begin{appendices}

\section{Ablations}

\subsection{Removing Skip-Cycles}
\label{appendix:a}

Removing the skip-cycle loss -- i.e. keeping only the long tracking cycle loss and dense similarity loss -- results in worse performance when applying the representations to the DAVIS-2017 dataset. This suggests the skip-cycle loss is useful in cases of occlusion or drift, and provides supplementary training data (c.f. Table \ref{tab:ablations1}).

\begin{table}[h]
\centering
\small
\vspace{-2mm}
\tablestyle{6pt}{1.05}
\begin{tabular}{l|c|c}
\multicolumn{1}{c|}{Experiment}  & $\mathcal{J}$(Mean) & $\mathcal{F}$(Mean) \\
\shline
Ours  &  {41.9} & {39.4} \\
\hline
Ours without Skip Cycles & 39.5 & 37.9 \\
\end{tabular}
\caption{Removing Skip Cycles, test on DAVIS.}
\label{tab:ablations1}
\end{table}

\subsection{Effect of k in k-NN Label Propagation}

We vary the number of nearest neighbors used in voting for label propagation (Eq.~\ref{eq:knn}), finding that aggregating fewer nearest neighbors improves performance (c.f. Table \ref{tab:ablations2}).

\begin{table}[h]
\centering
\small
\vspace{-2mm}
\tablestyle{6pt}{1.05}
\begin{tabular}{l|c|c}
\multicolumn{1}{c|}{Experiment}  & $\mathcal{J}$(Mean) & $\mathcal{F}$(Mean) \\
\shline
Ours (5-NN)&  {41.9} & {39.4} \\
\hline
Ours (20-NN) &  {40.8} & {38.5} \\
Ours (10-NN) &  {41.5} & {39.1} \\
Ours (1-NN) & {41.0} & {38.9} \\

\end{tabular}
\vspace{2mm}
\caption{Effect of $k$ in $k$-NN Label Propagation, test on DAVIS.}
\label{tab:ablations2}
\vspace{-4mm}
\end{table}

\subsection{Training with the Kinetics Dataset}
\label{appendix:kinetics}
Besides the VLOG dataset, we have also trained our model on the Kinetics Dataset~\cite{Kay2017}, which contains around 230K training videos (with 10s per video). Compared to the VLOG dataset, the Kinetics dataset contains more videos under less environment constraints: There are videos with both indoor and outdoor scenes; some videos also have large camera motion. After applying the learned representation for label propagation on DAVIS, we observe similar performance by training with VLOG and Kinetics datasets (c.f. Table \ref{tab:ablations4}).

\begin{table}[h]
\centering
\small
\vspace{-2mm}
\tablestyle{6pt}{1.05}
\begin{tabular}{l|c|c}
\multicolumn{1}{c|}{Experiment}  & $\mathcal{J}$(Mean) & $\mathcal{F}$(Mean) \\
\shline
Ours (VLOG)&  {41.9} & {39.4} \\
\hline
Ours (Kinetics) &  {42.5} & {39.2} \\
\end{tabular}
\vspace{2mm}
\caption{Train with VLOG / Kinetics, test on DAVIS.}
\label{tab:ablations4}
\vspace{-4mm}
\end{table}

\subsection{Fine-tuning on the Test Domain}
We emphasize that our method learns features that generalize even \textit{without} fine-tuning. Here we study the effect of fine-tuning on the DAVIS training set before testing. We find this does not improve test set performance significantly (c.f. Table~\ref{tab:ablations3}). There is a risk of overfitting since datasets like DAVIS are so small; this is part of the reason why unsupervised methods are desirable.   

\begin{table}[h]
\centering
\small
\vspace{-2mm}
\tablestyle{6pt}{1.05}
\begin{tabular}{l|c|c}
\multicolumn{1}{c|}{Experiment}  & $\mathcal{J}$(Mean) & $\mathcal{F}$(Mean) \\
\shline
Ours (ResNet-50)&  {41.9} & {39.4} \\
\hline
Fine-tune &  {42.0} & {39.1} \\
\end{tabular}
\vspace{2mm}
\caption{Finetuning on DAVIS train before test.}
\label{tab:ablations3}
\vspace{-4mm}
\end{table}

\section{Capacity of $\mathcal{T}$}
\label{appendix:b}
As mentioned in Section \ref{ssec:specs}, the tracking operation $T$ is deliberately constrained in capacity in order to maximize the representational responsibility of $\phi$. In our implementation, the only  parameters learned by $\mathcal{T}$ are those of the localizer $g$, which processes the affinity tensor $A$ to estimate the localization parameters. The affinity $A$ ($\mathbb{R}^{900 \times 100}$) is first reshaped to a tensor with dimension $\mathbb{R}^{900 \times 10 \times 10}$ as the input for $g$. The localizer $g$ is a small ConvNet with two convolutional layers ($3\times 3$ kernels with 512 channels) and one fully connected layer. The output of the ConvNet is a 3-dimension vector corresponding to 2D translation and rotation.

\section{Correspondence Visualization}
\label{appendix:c}

In Fig.~\ref{fig:flowResult} we visualize the correspondences (top-1 nearest neighbor) between regions with large movement in consecutive frames, comparing our features to ImageNet pre-trained features. 
Our method produces more detailed correspondence.  However, for certain object-level tasks (e.g. DAVIS), high-level semantics (captured by ImageNet) are more useful than good correspondences, which explains the difference in performance. 

\begin{figure}[h]
{
  \centering
  \includegraphics[width=1.0\columnwidth]{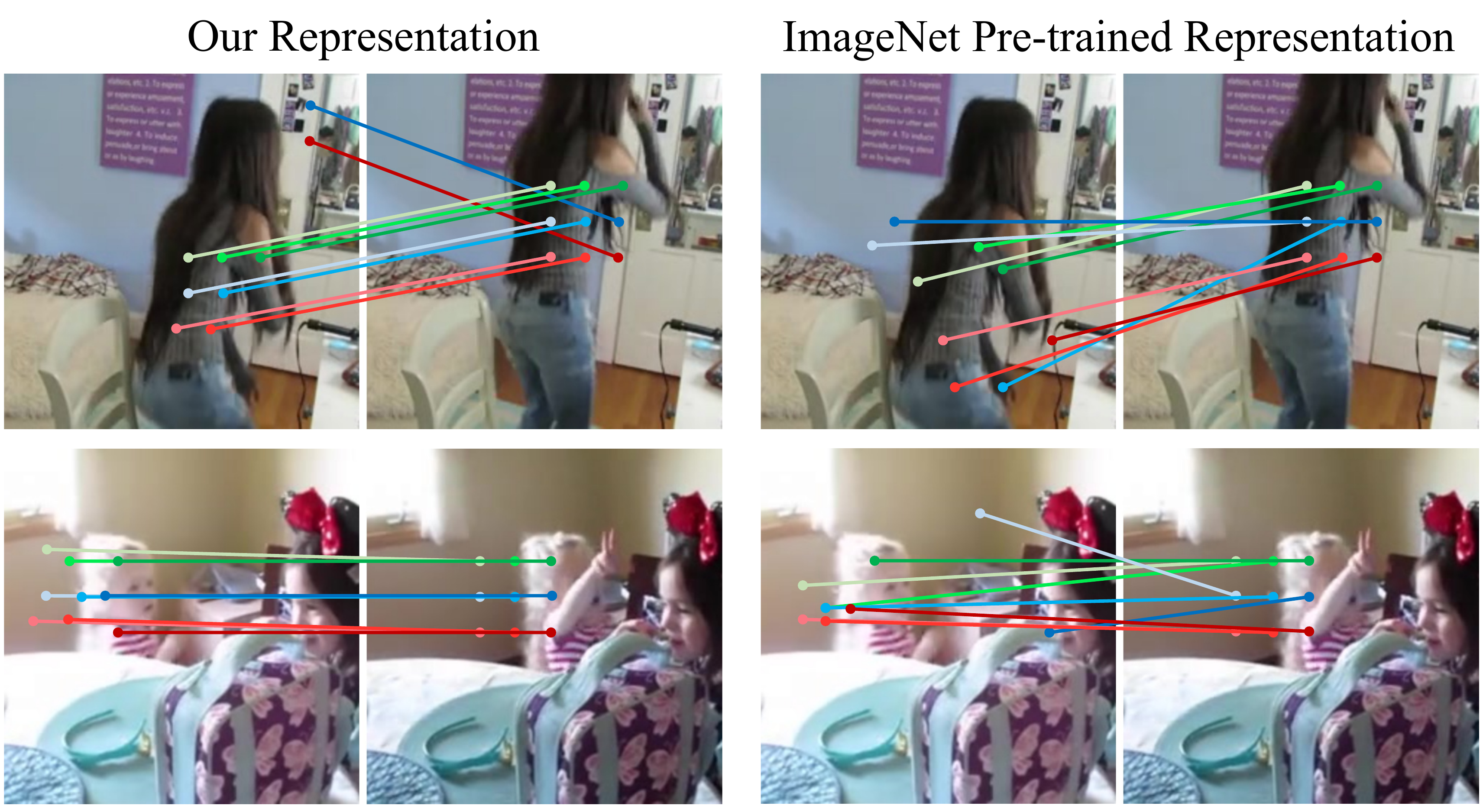}%
  \vspace{-0.1in}
  \caption{Visualizations of correspondence.}
  \label{fig:flowResult}
}
\end{figure}

\end{appendices}

\end{document}